\documentclass{article}

\usepackage[utf8]{inputenc}
\usepackage{spconf}
\usepackage{epstopdf}
\usepackage{amsmath}
\usepackage{amsfonts}
\usepackage{graphicx}
\usepackage{subcaption}
\usepackage{footnote}
\usepackage{float}
\usepackage{url}
\usepackage{color}

\graphicspath{{images/}}

\title{Learning an {MR} acquisition-invariant representation\\ using Siamese neural networks}
\name{Wouter M. Kouw$^{\dagger \star}$\thanks{WMK acknowledges support from the Niels Stensen Fellowship. AMM acknowledges support from ZonMw, IMDI Grant 104002002 (Brainbox).}\qquad Marco Loog$^{\star \dagger}$ \qquad Wilbert Bartels$^{*}$ \qquad Adriënne M. Mendrik$^{* +}$}

\address{$^{\star}$Delft University of Technology, $^{\dagger}$University of Copenhagen, \\ $^{*}$University Medical Center Utrecht, $^{+}$Netherlands eScience Center}
\begin{document}
%
\maketitle
\begin{abstract}
Generalization of voxelwise classifiers is hampered by differences between MRI-scanners, e.g. different acquisition protocols and field strengths. To address this limitation, we propose a Siamese neural network ({\sc mrai-net}) that extracts acquisition-invariant feature vectors. These can consequently be used by task-specific methods, such as voxelwise classifiers for tissue segmentation. {\sc mrai-net} is tested on both simulated and real patient data. Experiments show that {\sc mrai-net} outperforms voxelwise classifiers trained on the source or target scanner data when a small number of labeled samples is available.

\end{abstract}
\section{Introduction} \label{sec:intro}
Voxelwise classifiers for brain tissue segmentation should be trained on a sufficiently large representative data set, covering all possible types of variation. However, acquiring manual labels as ground truth is both labor intensive and time consuming. Furthermore, non-standardized manual segmentation protocols and inter- and intra-observer variability add another factor of variation to an already complex problem. Instead of increasing the number of manual labels, we propose to improve generalization by teaching a neural network to minimize an undesirable form of variation, namely acquisition-based variation. The proposed network learns a representation \cite{bengio2013RepLearning} in which for example gray matter patches acquired with a 1.5T scanner and a 3T scanner are considered similar. Therefore it has the potential to fully exploit a 1.5T data set with fully labeled brain tissues for segmenting an unlabelled 3T data set.


Overcoming acquisition-variation is a relatively new challenge in medical imaging. Transfer classifiers have been proposed that focus on weighting classifiers based on how well their training data matches the test data, such as weighted SVM's \cite{van2015transfer} and weighted ensembles \cite{cheplygina2016asymmetric}. However, these classifiers need to be retrained for every new test data set, and do not remove acquisition-variation in general or extract acquisition-invariant feature vectors for later use by task-specific methods.

We propose to learn a task-independent representation, in which acquisition variation is minimal while tissue variation is maintained. Patches sampled from MRI-scans that are mapped to this new representation will become feature vectors, and can be used by task-specific classifiers later on. In order to minimize one factor of variation while maintaining another, we exploit a Siamese network \cite{hadsell2006dimensionality}. The proposed network ({\sc mrai-net}) is described in Section \ref{method}. Experiments on both simulated and real patient data are shown in Section \ref{experiment}.

\section{MR acquisition-invariant network}\label{method}
Suppose that we have scans that are acquired in two different ways; A (source) and B (target). A tissue patch, e.g. gray matter, is selected from both scans A and B. The aim is to teach a neural network that both these patches are gray matter. To achieve this, we use a loss function that expresses that pairs of samples from the same tissue but different scanners should be \emph{similar}. However, if the neural network only receives this expression it would map all patches to a single point and would destroy variation between tissues. To balance out the action of making certain pairs more similar, another expression is added, stating that patches from different tissues -- regardless of scanner -- should remain \emph{dissimilar}.

\subsection{Siamese loss} \label{sec:siam_loss}
Neural networks transform data in each layer. We summarize the total transformation from input to output with the symbol $f$, i.e. patch $a$ will be mapped to the new representation with $f(a)$ and patch $b$ will be mapped with $f(b)$. Distance in the new representation is expressed as $d_{f}(a,b) = \| f(a) - f(b)\|$, where $\| \cdot \|$ is an $L^1$-norm. Pairs marked as similar (y=1) should be pulled together, while those marked as dissimilar (y=0) should be pushed apart. The loss for the similar pairs consists of the squared distance, $\ell_{\text{sim}}(f \mid a,b) = d_{f}(a,b)^2$. The loss function for the dissimilar pairs consists of a hinge loss: $\ell_{\text{dis}}(f \mid a,b) = \max \big [0, m- d_f(a,b) \big]$ where $m$ is the margin parameter. Pairs that are pushed past the margin, will not suffer a loss.

We can combine the similar and dissimilar losses into a single loss function:
\begin{align}
	\ell(f) =& \sum_{i} \ y_i \ \ell_{\text{sim}}\left(f \mid a_i,b_i \right) + (1-y_i) \ \ell_{\text{dis}} \left( f \mid a_i,b_i \right) \nonumber \\
		   =& \sum_{i} \ y_i \ d_f(a_i, b_i)^2 + (1-y_i) \max \left[ 0, m - d_{f}(a_i, b_i) \right] \, . \nonumber
\end{align}
where $i$ iterates over pairs. This type of loss function is known as a \emph{Siamese} loss \cite{hadsell2006dimensionality}.

\subsection{Labeling pairs as similar or dissimilar} \label{sec:sim_labeling}
Assume that we have sufficient manual segmentations (voxel labels) on scans from scanner A to train a supervised classifier, but a limited amount of labels from scanner B.  Let $K$ be the set of tissue labels. A patch from scanner A is denoted $a_{t}$, and a patch from scanner B is denoted $b_{t}$, with $t$ specifying the current patch's tissue. Given sets of patches, we form similar and dissimilar pairs, designated by a similarity label $y$. The following pairs are labeled as similar ($y=1$): source patches from the same tissue ($k \in K$) $\{(a_{t=k},a_{t=k})\}$, source and target patches from the same tissue: $\{(a_{t=k},b_{t=k})\}$, and target patches from the same tissue: $\{(b_{t=k},b_{t=k})\}$. Conversely, the following are labeled as dissimilar ($y=0$): source patches from different tissues $(k,l \in K : k \neq l)$ $\{(a_{t=k},a_{t=l})\}$, source and target patches from different tissues $\{(a_{t=k},b_{t=l})\}$, and target patches from different tissues $\{(b_{t=k},b_{t=l})\}$. 

\begin{figure}[htb]
\centering
\includegraphics[width=.45\textwidth]{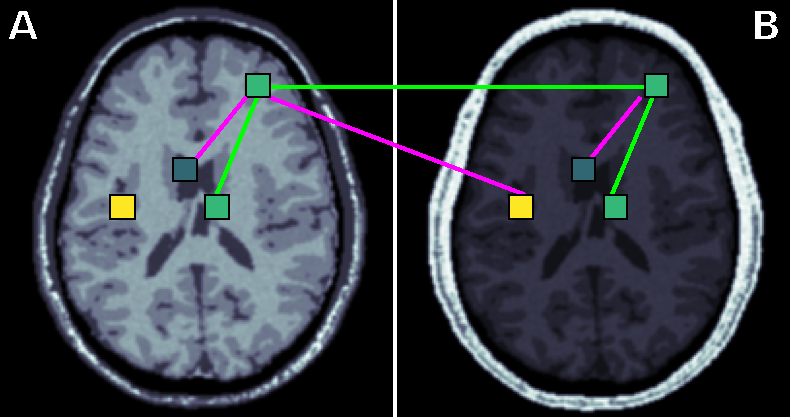}
\caption{Illustration of extracting pairs of patches from images from scanner A and B. Each image shows 4 patches: 2 gray matter ones (green), 1 cerebrospinal fluid (blue) and 1 white matter (yellow). The lines mark the 6 types of combinations from Section \ref{sec:sim_labeling} (green = similar, purple = dissimilar).}
\label{fig:sim_labeling}
\end{figure}

Let $N_k$ be the number of patches extracted from a scan of scanner $A$ belonging to tissue $k$, and $M_k$ be the number of patches extracted from scanner $B$ of tissue $k$. In total, the number of combinations is $\sum_{k \in K} (N_k + M_k)^2 + \sum_{(k,l) \in {K \choose 2}} (N_k N_{l} + N_{k} M_{l} + M_{k} M_{l})$, where $(k,l) \in {K \choose 2}$ refers to all combinations of $2$ that can be taken from the set of tissues. The combinatorial explosion works in our favor, as it allows us to generate a large training data set from only a few labeled target samples. Figure \ref{fig:sim_labeling} illustrates the process of selecting pairs of patches from different scanners.

\subsection{Network architecture} \label{sec:net_arch}
The network consists of two pipelines and a Siamese loss layer that acts on the pipes' output layers. We made the following architectural choices: 15x15 input patches, 8 convolution kernels of size 3x3 with "ReLU" activation functions, a fully-connected layer of size 16, another fully-connected layer of size 8, and a final fully-connected layer of size 2. Dropout was set to 0.2 during training, and we used a standard "RMSprop" optimizer to perform backpropagation. For more implementation details, see the accompanying software repository: \url{github.com/wmkouw/mrai-net}. {\sc mrai-net} is implemented in a combination of Tensorflow and Keras.

Patches represented in the final representation layer are, in fact, feature vectors. The wider the layer, the higher the feature vector dimensionality. The two pipelines share their weights, which means they are constrained to perform the same transformation. This means that \emph{single patches} can be fed through the network and that it is not necessary to form pairs at test time.

\section{Experiment} \label{experiment}
In this experiment we test the proxy ${\cal A}$-distance between patches from the source and target scanners and we compare the performance of a linear classifier trained on {\sc mrai-net}'s feature vectors on a cross-scanner tissue segmentation task.

\subsection{Data}\label{sec:data}
We simulated different MR acquisitions from anatomical models of the human brain \cite{aubert2006twenty}, using the MRI simulator SIMRI \cite{benoit2005simri, aubert2006twenty}. The anatomical models consist of transverse slices of 20 normal brains (Brainweb). We simulated two acquisition types: (1) Brainweb1.5T, a standard gradient-echo acquisition protocol with the same parameters as the MRI-scanner in the Rotterdam Scan Study (B0 = 1.5T, $\theta=20^{\circ}$, TR=13.8 ms, TE=2.8 ms) \cite{ikram2015rotterdam}, and (2) Brainweb3.0T, a standard gradient-echo protocol with the same parameters as the scanner used for MRBrainS (B0 = 3.0T, $\theta=90^{\circ}$, TR=7.9 ms, TE=4.5 ms) \cite{mendrik2015mrbrains}. Magnetic field inhomogeneities and partial volume effects are not included in the simulation. There are 9 tissues, but we grouped these into "background", "cerebrospinal fluid", "gray matter", and "white matter". The simulations result in images of 256 by 256 pixels, with a 1.0x1.0mm resolution. Figure \ref{fig:sim_labeling} shows examples of the Brainweb1.5T (A) and Brainweb3.0T (B) scan of the same subject.

In order to test the proposed method on real data, we use the publicly available training data (5 subjects) from the MRBrainS challenge \cite{mendrik2015mrbrains}.

\subsection{Measuring acquisition variation}\label{sec:MRAImeasure}
The proxy ${\cal A}$-distance is a measure of the discrepancy between two data sets \cite{ben2010theory}. Denoted by $d_{\cal A}$, it is defined as: $d_{\cal A}(a, b) = 2(1 - 2 e(a, b))$, where $e$ represents the test error of a classifier trained to discriminate patches $a$ from scanner A and patches $b$ from scanner $B$. For computing the proxy ${\cal A}$-distance, we draw 1500 patches from all source and 1500 from all target scans. A linear support vector machine is trained to discriminate between them, and the cross-validation error is used to produce $e(a,b)$.

\subsection{Measuring tissue variation}\label{sec:TissuePreservation}
A tissue classifier is used to measure how much variation between tissues is preserved in {\sc mrai-net}'s representation, specifically gray matter, white matter and cerebrospinal fluid. For evaluation, we use scans from target subjects that have been held back (10 subjects). From these scans, we draw 50 patches per tissue at random, for a total of 1500 patches. We apply the tissue classifier to these test samples and compute the classification error rate. 

\subsection{Experimental setup}
Ultimately, we know that tissue variation is preserved if the extracted feature vectors can be used for tissue segmentation. To that end, we compare a linear support vector machine trained on {\sc mrai-net}'s extracted feature vectors (also referred to as {\sc mrai-net}) to two other supervised classifiers: (1) {\sc source} classifier, a convolutional neural network (CNN) trained on samples from the source (4 subjects) and target data (1 subject), and (2) {\sc target} classifier, a CNN trained on samples from the target data (1 subject). These two classifiers represent two possible scenario's where you would \emph{not} account for the differences between the scanners. {\sc source} and {\sc target}'s network architecture is the same as that of each pipeline in {\sc mrai-net}; this rules out that differences in behavior between {\sc source}, {\sc target} and {\sc mrai-net} are due to choices for specific architectures. All classifiers are trained on a range between 1 and 1000 labeled target patches per tissue.

We first performed this experiment using Brainweb1.5T as the source scanner and Brainweb3T as the target scanner. Since the same subjects are used, all variation between the data sets is acquisition-based. Secondly, we performed the same experiment using Brainweb1.5T as the source scanner and MRBrainS as the target scanner. Now there are more factors of variation, such as different subjects, environments, partial volume effects and field inhomogeneities.

\begin{figure}[htb]
\centering
\includegraphics[width=.49\textwidth]{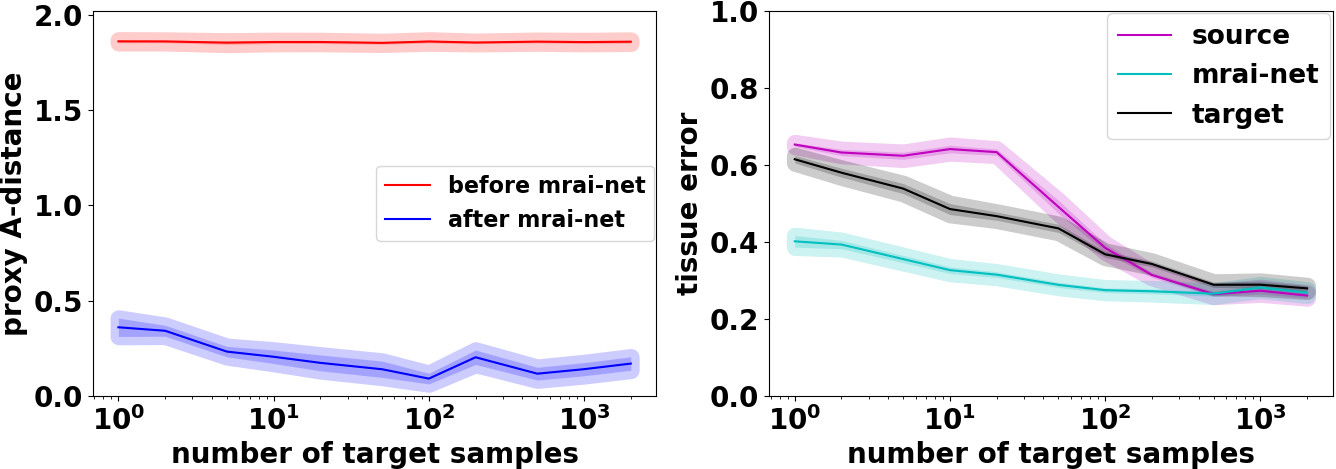}\\
\includegraphics[width=.49\textwidth]{err-lc_exp_acqinv_cnn_b1mb_H.png}
\caption{Learning curves for Brainweb1.5T $\rightarrow$ Brainweb3T (Top row) and Brainweb1.5T $\rightarrow$ MRBrainS (Bottom row). (Left column) Proxy A-distance between source and target patches before (red) and after (blue) learning the new representation (smaller is better). (Right column) Tissue classification error for {\sc source}, {\sc mrai-net} and {\sc target}. 
}
\label{fig:results}
\end{figure}

\subsection{Results} \label{sec:MultSamples}
Figure \ref{fig:results} shows the proxy ${\cal A}$-distance and the tissue classification error, with an increasing number of labeled target patches used for training. In general, the experiment on the real data (MRBrainS) follows the same pattern as the simulated data. By using {\sc mrai-net}, the distance between the source and target scanner data sets (proxy ${\cal A}$-distance) drops substantially, even with only one labeled target sample per class. With one hundred target training samples the proxy ${\cal A}$-distance approaches $0$ (small acquisition variation means the data sets overlap), while tissue variation is preserved (tissue classification error 0.11 for simulated data and 0.27 for MRBrainS real patient data). The tissue classification error for the {\sc source} and {\sc target} voxel classifiers is 0.21 and $~$0.37, respectively.    

\begin{figure*}[htb]
\centering
\begin{subfigure}[t]{.18\textwidth}
	\includegraphics[width=.98\textwidth]{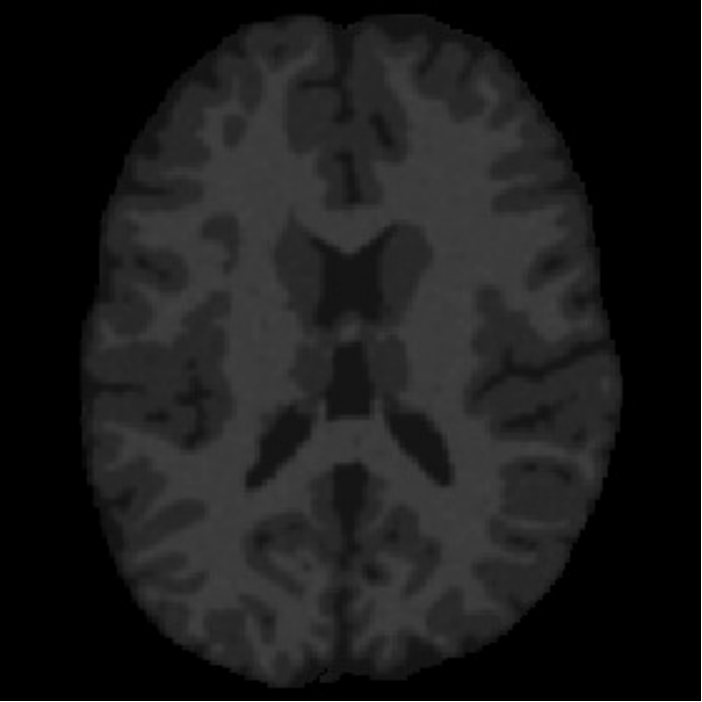}
    \caption{Scan}
\end{subfigure} \ 
\begin{subfigure}[t]{.18\textwidth}
	\includegraphics[width=.98\textwidth]{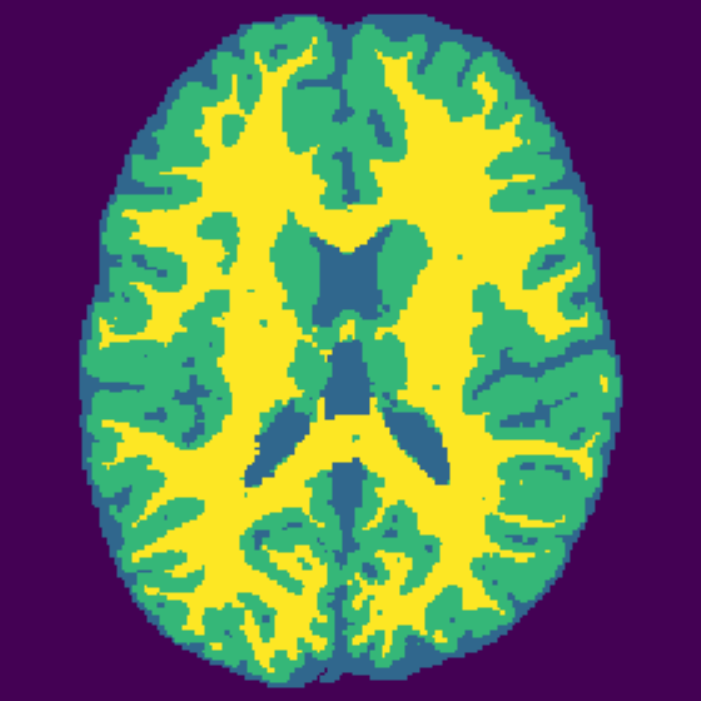}
    \caption{Ground truth}
\end{subfigure} \
\begin{subfigure}[t]{.18\textwidth}
	\includegraphics[width=.98\textwidth]{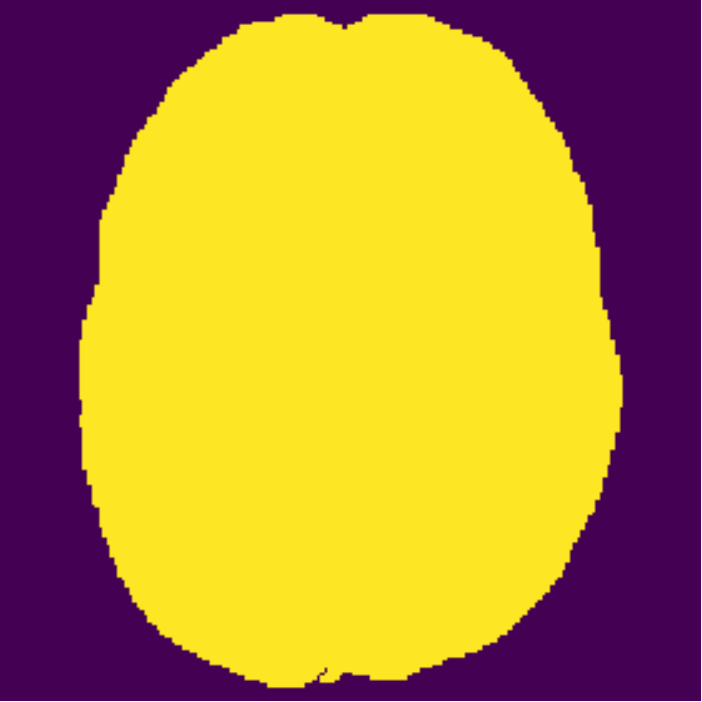}
    \caption{{\sc source}}
\end{subfigure} \ 
\begin{subfigure}[t]{.18\textwidth}
	\includegraphics[width=.98\textwidth]{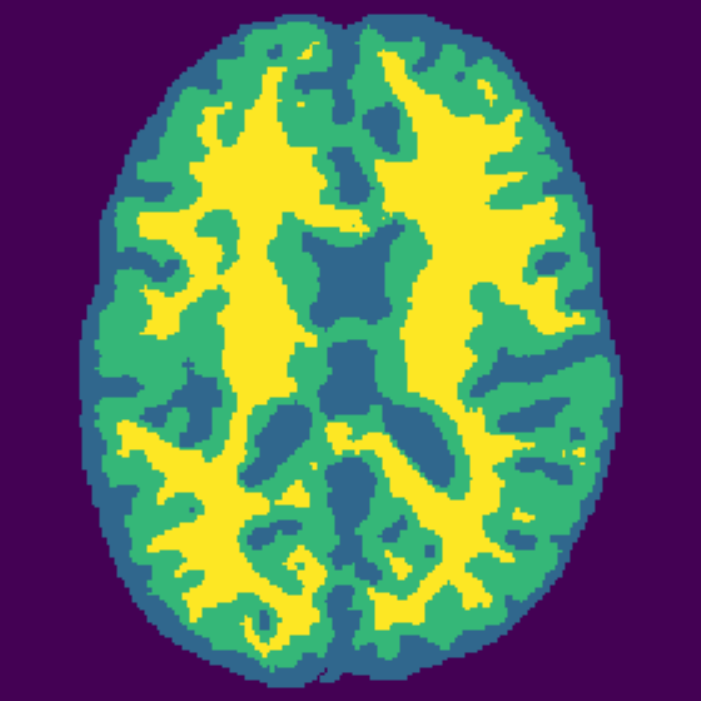}
    \caption{{\sc mrai-net}}
\end{subfigure} \ 
\begin{subfigure}[t]{.18\textwidth}
	\includegraphics[width=.98\textwidth]{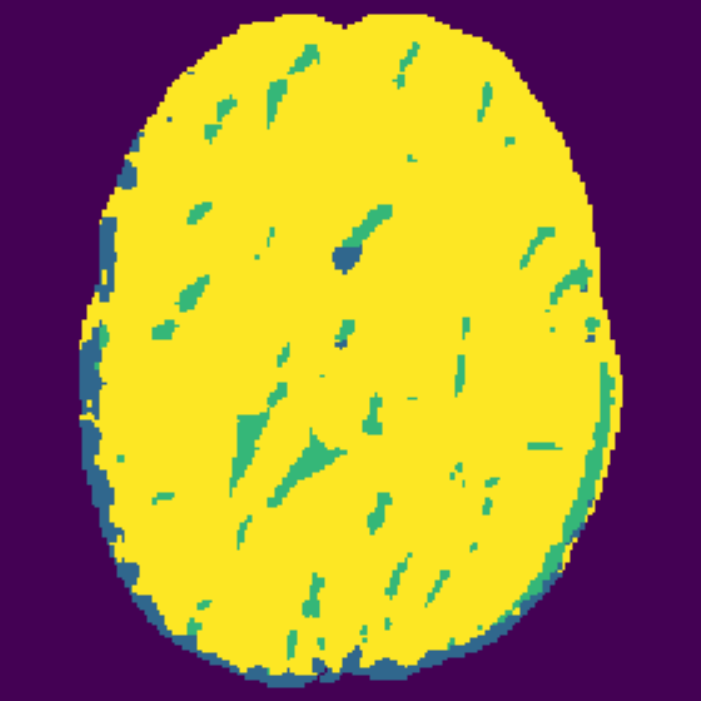}
    \caption{{\sc target}}
\end{subfigure} \\
\begin{subfigure}[t]{.18\textwidth}
	\includegraphics[width=.98\textwidth]{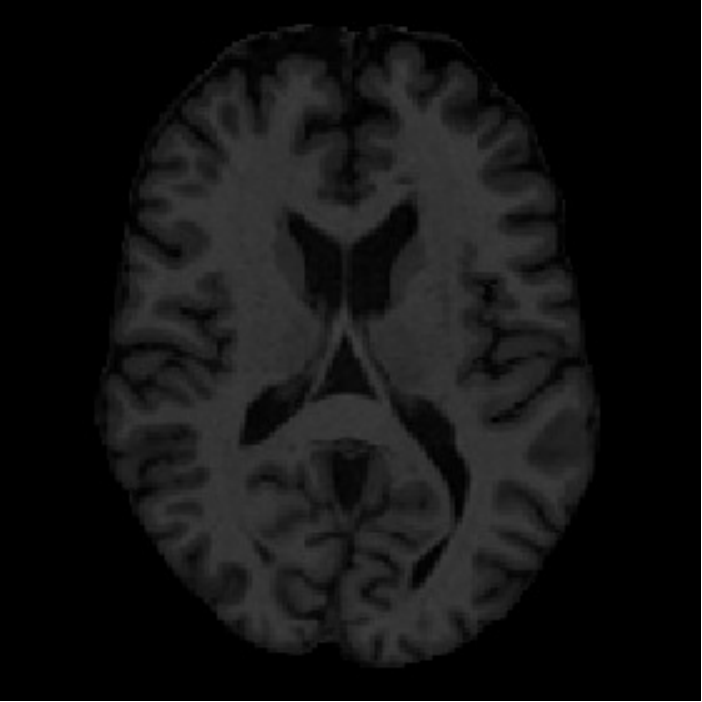}
    \caption{Scan}
\end{subfigure} \
\begin{subfigure}[t]{.18\textwidth}
	\includegraphics[width=.98\textwidth]{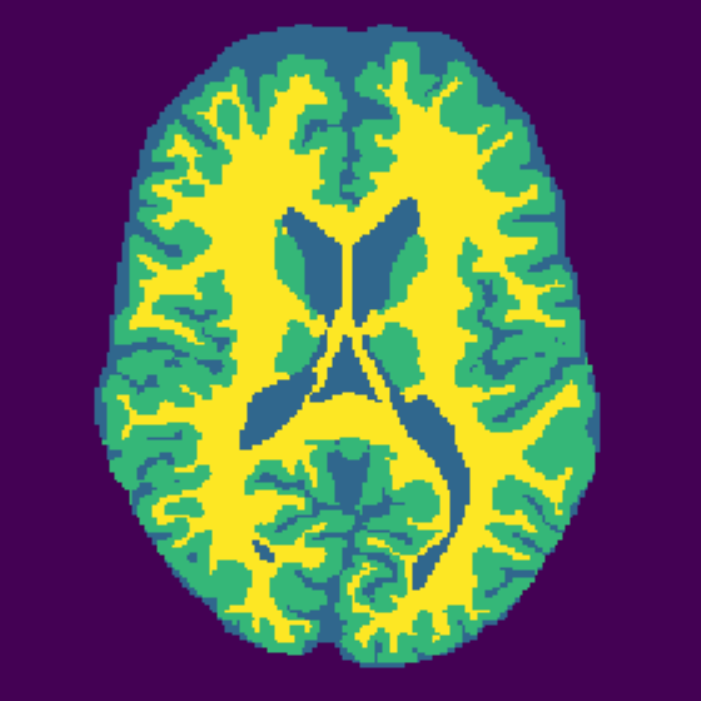}
    \caption{Ground truth}
\end{subfigure} \
\begin{subfigure}[t]{.18\textwidth}
	\includegraphics[width=.98\textwidth]{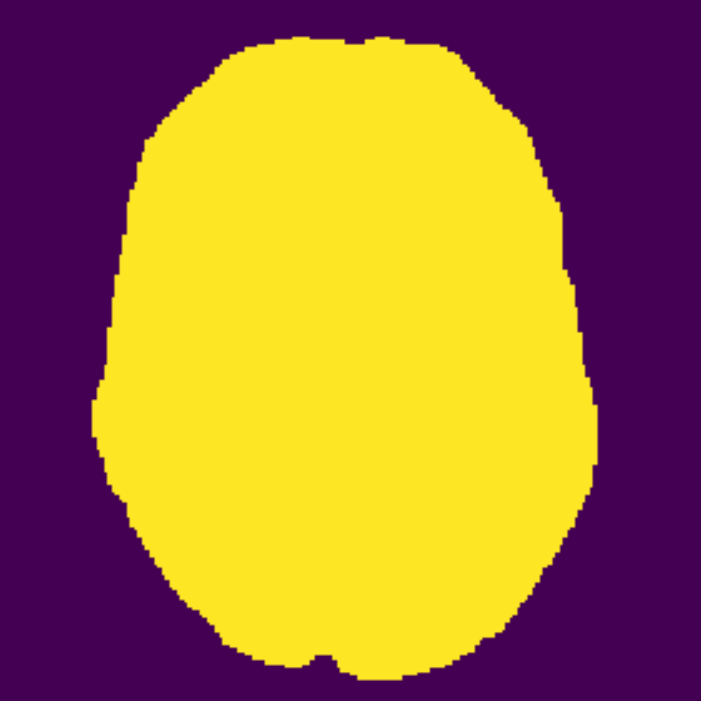}
    \caption{{\sc source}}
\end{subfigure} \ 
\begin{subfigure}[t]{.18\textwidth}
	\includegraphics[width=.98\textwidth]{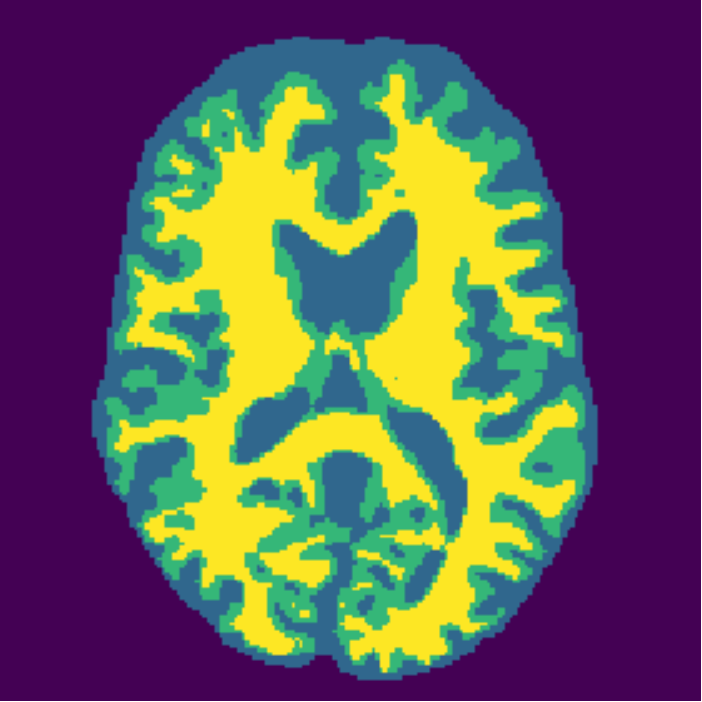}
    \caption{{\sc mrai-net}}
\end{subfigure} \
\begin{subfigure}[t]{.18\textwidth}
	\includegraphics[width=.98\textwidth]{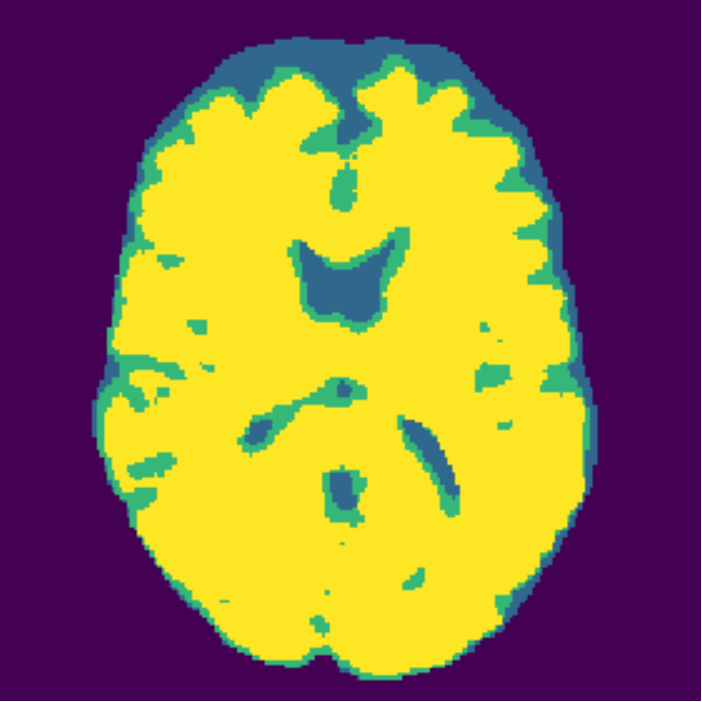}
    \caption{{\sc target}}
\end{subfigure}
\caption{Example segmentations into white matter (yellow), gray matter (green) and cerebrospinal fluid (blue) using only one labeled target patch per class, for Brainweb1.5T $\rightarrow$ Brainweb3T (top row) and Brainweb1.5T $\rightarrow$ MRBrainS (bottom row).}
\label{fig:preds_b1b3}
\end{figure*}

For ten labeled target sample per tissue, {\sc mrai-net}'s error is 0.17 (simulated data) and 0.33 (MRBrainS data), while {\sc source} still performs at a 0.66/0.64 error (simulated/MRBrainS) and {\sc target} performs at 0.40/0.49. Given sufficient samples, all three classifier reach similar performances. Figure \ref{fig:preds_b1b3} illustrates the difference in tissue classification performance when only one labeled target sample per tissue is used for training.





Note furthermore that {\sc source} shows worse performance than {\sc target} for less than roughly 50 samples. In this setting, the scanners are so different that including the {\sc source} samples in the training set actually interferes with learning. Given enough target samples, however, {\sc source} finds a good balance between source and target samples and starts to match the performance of {\sc target}. 

\section{Conclusion}
We proposed a Siamese neural network ({\sc mrai-net}) to learn a representation of the data where acquisition-based variation is minimal and tissue-based variation is maintained. A linear classifier trained on feature vectors extracted by {\sc mrai-net} outperforms conventional CNN classifiers trained on the source and target data sets on a cross-scanner tissue segmentation task, when few labeled target samples are available.


\bibliographystyle{IEEEbib}
\bibliography{kouw_isbi19a}

\end{document}